\documentclass{article} % For LaTeX2e
\usepackage{iclr2024_conference,times}

\usepackage{amsmath,amsfonts,bm}

\def\Figref#1{Figure~\ref{#1}}

\def\Secref#1{Section~\ref{#1}}
\def\eqref#1{equation~\ref{#1}}
\def\Eqref#1{Equation~\ref{#1}}
\def\1{\bm{1}}

\DeclareMathAlphabet{\mathsfit}{\encodingdefault}{\sfdefault}{m}{sl}
\SetMathAlphabet{\mathsfit}{bold}{\encodingdefault}{\sfdefault}{bx}{n}

\newcommand{\E}{\mathbb{E}}

\newcommand{\Var}{\mathrm{Var}}

\newcommand{\Cov}{\mathrm{Cov}}
\usepackage{hyperref}
\usepackage{url}

\usepackage{xspace}
\usepackage{graphicx}
\usepackage{wrapfig}
\usepackage{float}
\usepackage{subfigure}
\usepackage{xfrac}

\newcommand{\eg}{\emph{e.g.},\xspace}
\newcommand{\ie}{\emph{i.e.},\xspace}

\usepackage{amsthm}
\usepackage{amssymb}
\newtheorem{theorem}{Theorem}
\newtheorem{apptheorem}{Theorem}
\newtheorem*{theorem*}{Theorem}
\newtheorem{lemma}{Lemma}

\newtheorem{definition}{Definition}

\newcommand{\ha}{\mathcal H_{\mathcal A}}
\newcommand{\oE}{\mathop\E}
\newcommand{\oVar}{\mathop\Var}
\newcommand{\oCov}{\mathop\Cov}
\newcommand{\err}{\mathrm{err}}

\title{On the Variance of Neural Network Training with respect to Test Sets and Distributions}

\author{
Keller Jordan\thanks{Experiments performed while at Hive AI} \\
\small{\texttt{kjordan4077@gmail.com}}
}

\iclrfinalcopy % Uncomment for camera-ready version, but NOT for submission.
\begin{document}

\maketitle

\vspace{-2mm}
\begin{abstract}
\vspace{-2mm}
Typical neural network trainings have substantial variance in test-set performance between repeated runs, impeding hyperparameter comparison and training reproducibility.
In this work we present the following results towards understanding this variation.
(1) Despite having significant variance on their test-\textit{sets}, we demonstrate that standard CIFAR-10 and ImageNet trainings have little variance in performance on the underlying test-\textit{distributions} from which their test-sets are sampled.
(2) We show that these trainings make approximately independent errors on their test-sets.
That is, the event that a trained network makes an error on one particular example does not affect its chances of making errors on other examples, relative to their average rates over repeated runs of training with the same hyperparameters.
(3) We prove that the variance of neural network trainings on their test-sets is a downstream consequence of the class-calibration property discovered by \citet{jiang2021assessing}. Our analysis yields a simple formula which accurately predicts variance for the binary classification case.
(4) We conduct preliminary studies of data augmentation, learning rate, finetuning instability and distribution-shift through the lens of variance between runs.
\end{abstract}

\vspace{-2mm}
\section{Introduction}
\vspace{-2mm}
Modern neural networks \citep{krizhevsky2017imagenet, he2016deep, vaswani2017attention} are trained using stochastic gradient-based algorithms, involving randomized weight initialization, data ordering, and data augmentations. Because of this stochasticity, each independent run of training produces a different network with better or worse performance than average.

This variance between such independent runs is often substantial. \citet{picard2021torch} shows that for standard CIFAR-10~\citep{cifar100} training configurations, there exist random seeds which differ by 1.3\% in terms of test-set accuracy. In comparison, the gap between the top two methods competing for state-of-the-art on CIFAR-10 has been less than 1\% throughout the majority of the benchmark's lifetime\footnotemark.
% \footnote{\url{https://paperswithcode.com/sota/image-classification-on-cifar-10}}.
Prior works therefore view this variance as an obstacle which impedes comparisons between training configurations~\citep{bouthillier2021accounting,picard2021torch} and reproducibility~\citep{bhojanapalli2021reproducibility,zhuang2022randomness}. To mitigate stochasticity, \citet{zhuang2022randomness} study deterministic tooling, \citet{bhojanapalli2021reproducibility} develop regularization methods, and many recent works~\citep{wightman2021resnet,liu2022convnet} report the average of validation metrics across multiple runs when comparing training configurations.

% In this paper we contribute new insights regarding the nature of variation between runs of neural network training.

% In this paper, we begin with the following two questions regarding the variance in test-set accuracy between independent, identically-configured runs of neural network training:
% \begin{enumerate}
%     % \item What causes this variance? Can we isolate which of source of randomness is most responsible?
%     \item If we train many times, and take the network which performed best on the test-\textit{set}, should we also expect it to perform above average on the test-\textit{distribution}?
%     \item Which hyperparameters affect variance? For example, do there exist training configurations with the same average accuracy, but different amounts of variance?
% \end{enumerate}
\footnotetext{\url{https://paperswithcode.com/sota/image-classification-on-cifar-10}}

In this work we contribute new results towards understanding the variance between runs of neural network trainings. We use an empirical approach involving hundreds of thousands of trained models in order to answer the following questions.

% In this paper we investigate the following questions about variance between runs of neural network training. We use an empirical approach involving hundreds of thousands of trained models.
% In this paper we do not try to mitigate variance, but rather to simply understand it. We use an empirical approach involving hundreds of thousands of trained models in order to investigate the following questions.
\begin{enumerate}
    \item Repeatedly running a standard training yields a series of models with often substantial variance in test-set accuracy. Do such models have genuine differences in underlying quality? Or is this variance just a form of finite-sample noise due to the limited size of the test-set?
    \item Does the empirical distribution of test-set accuracy across repeated runs of training possess any definable structure?
    \item Is there any way to estimate the variance of a given training configuration a priori, \ie without the need to empirically measure it across many repeated runs?
\end{enumerate}
Both the first and last of these questions have immediate practical consequences.
For the first question, if the answer is that every run of training yields a model with the same underlying performance on held-out batches of test data, then practitioners can confidently execute just a single run per configuration. Otherwise, a superior strategy would be to take the best model from multiple trials.
% If the answer to the first question is that every run of training yields a model with the same underlying quality, \ie the same performance on held-out batches of test data, then practitioners can confidently execute just a single run per configuration. Otherwise, a superior strategy would be to take the best model from multiple trials.
For the third question, if a method for estimating variance a priori does exist, then this provides a useful tool by which practitioners can confidently estimate the statistical significance of hyperparameter comparisons, without the need for many runs of training.
To answer all three questions, we contribute the following results.
% And if a method for estimating variance a priori does exist, as the third question suggests, then this provides a useful tool by which practitioners can confidently estimate the statistical significance of hyperparameter comparisons without the need for many runs of training.
% To answer all three questions, we contribute the following results.
% \vspace{-1mm}
\begin{itemize}
    \item Random seeds which are ``lucky'' with respect to one set of test data perform no better than average with respect to a second set. (\Secref{sec:do_lucky})
    \item Over repeated runs of training, the distribution of errors made by trained networks can be approximately explained via the framework of \textit{independent errors}. (\Secref{sec:hypothesis})
    \item Although standard trainings have substantial variance on their test-sets, they have little variance in performance on their underlying test-distributions. (\Secref{sec:estimating_distrib})
    \item Variable test-set performance is a downstream effect of the class-calibration property~\citep{jiang2021assessing} of neural network trainings. For the binary classification case, this property implies a simple formula which accurately predicts variance a priori. (\Secref{sec:ece_bound})
\end{itemize}
% \vspace{-1mm}

Our experiments show that these results hold true for standard training configurations across both CIFAR-10 and ImageNet~\citep{deng2009imagenet}. As a limitation, we show that they do \textit{not} hold true for two exceptional scenarios:
trainings with pathological instability~(\Secref{sec:bert_finetune}), and trainings where there is a shift between the training and test distributions~(\Secref{sec:shift}, \Secref{sec:imagenet}). Both of these cases have a large distribution-wise variance, differing from our results in the standard cases.

To complete our study of variance, 
we additionally conduct preliminary investigations regarding the effect of learning rate~(\Secref{sec:learning_rate}) and data augmentation~(\Secref{sec:data_augs}). We find that when increasing the learning rate, accuracy begins to decline at the same point at which significant distribution-wise variance appears. And we find that data augmentation reduces variance, although the mechanism by which this happens is not yet clear.

\vspace{-1mm}
\subsection{Related work}
\vspace{-1mm}
A number of prior works investigate which sources of stochasticity are most responsible for the variation between runs of training. \citet{fort2019deep} observe that when using a below-optimal learning rate, randomized data ordering has a smaller impact than model initialization on the churn of predictions between runs. \citet{bhojanapalli2021reproducibility} similarly find that fixing the data ordering has no effect, while fixing the model initialization reduces churn. On the other hand, \citet{bouthillier2021accounting} report that data ordering has a larger impact than model initialization. And finally, \citet{summers2021nondeterminism} find instead that most variation can be attributed to the high sensitivity of the training process to initial conditions, by showing that a single bit difference in starting parameters leads to the full quantity of prediction churn between runs. We replicate the results of \citet{summers2021nondeterminism} in \Secref{sec:factors}, although we find that the result depends upon the training duration.

\citet{dodge2020fine} study variation between runs of BERT$_\text{LARGE}$ finetuning, and achieve substantial gains in validation performance via the strategy of re-running finetuning many times and taking the best-performing result.
We demonstrate (\Secref{sec:bert_finetune}) that for the case of BERT$_\text{BASE}$, the low amount of genuine distribution-wise variance between runs indicates that any performance gains yielded by this strategy would only amount to overfitting the validation set. On the other hand, for BERT$_\text{LARGE}$ we demonstrate that there is genuinely significant distribution-wise variance, supporting the use of multiple runs of training as suggested. \citet{mosbach2020stability} also study the finetuning instability of BERT$_\text{LARGE}$, and suggest that it can be mitigated by warming up the learning rate, training for longer with a smaller learning rate, and using bias correction for Adam~\citep{kingma2014adam}.

Earlier works on neural network ensembles have found that they are more well-calibrated than individual networks~\citep{lakshminarayanan2017simple,nixon2020bootstrapped}. And \citet{mukhoti2021deep} observed that the usefulness of an ensemble's uncertainty scores depends upon variance between the individual networks.
Our theoretical results in \Secref{sec:ece_bound} draw upon the related \textit{class-wise calibration property} of neural network trainings. This property was discovered by \citet{jiang2021assessing}, who used it to obtain a theoretical proof of the empirical phenomenon that the disagreement rate between two independently trained networks is approximately equal to their error rates~\citep{nakkiran2020distributional,jiang2021assessing}.

\begin{figure}
    \centering
    \includegraphics[width=0.95\textwidth]{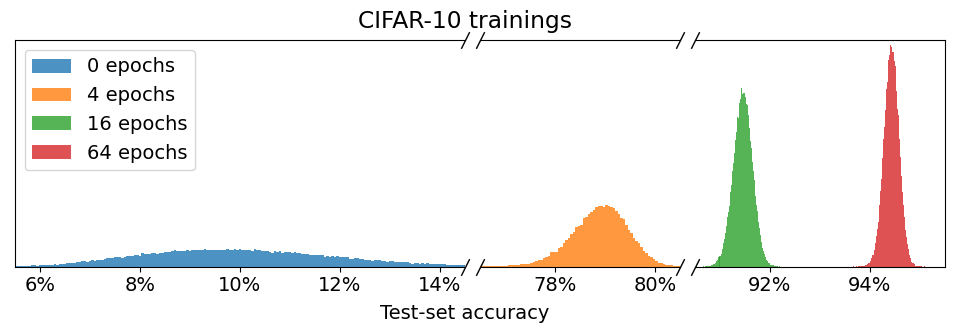}
    \vspace{-2mm}
    \caption{\small \textbf{Accuracy distributions.} The test-set accuracy distributions across our four training durations, displayed as unsmoothed histograms for 60,000 repeated runs of training each. The differences between the ``luckiest'' and most unlucky run (max minus min accuracy) are 13.2\%, 6.6\%, 1.7\%, and 1.4\% for the 0, 4, 16, and 64-epoch training durations, respectively. The standard deviations are 1.87\%, 0.56\%, 0.19\%, and 0.15\%.}
    \label{fig:distributions}
\vspace{-4mm}
\end{figure}

Several prior works~\citep{baldock2021deep,ilyas2022datamodels,lin2022measuring} study the effect of randomly varying the data used to train a neural network, across a large number of training runs.
Our study differs from these works in that we consider the simpler scenario of running a single training algorithm many times without varying anything except the random seed.

Broadly, our work is related to research aiming to understand the relationship between pairs of neural networks produced by repeated runs of training. This topic is of both theoretical and practical interest, and has been studied from a variety of angles, including the similarity of internal representations \citep{li2015convergent,kornblith2019similarity}, degree of correlation between predictions \citep{fort2019deep,jiang2021assessing}, similarity of decision boundaries~\citep{somepalli2022can}, path-connectivity in weight-space \citep{draxler2018essentially,garipov2018loss}, and linear mode connectivity \citep{frankle2020linear,tatro2020optimizing,entezari2021role}.

\vspace{-1mm}
\section{Setup}
\vspace{-2mm}
\label{sec:setup}
\textbf{Notation.} We study supervised classification problems in which test-set examples are sampled independently from a distribution $\mathcal D$ over $\mathcal X \times \mathcal Y$, where $\mathcal Y = \{1, \dots, k\}$ is the set of classes and $\mathcal X$ is the input space. We make no assumptions on the training distribution.
By a \textit{training algorithm} or configuration, we mean a training pipeline which includes everything necessary for training besides the random seed. That is, we assume training algorithms already include the choice of optimization algorithm, dataset, network architecture, and hyperparameters, so that only the random seed remains to determine the outcome of training.
Following \citet{jiang2021assessing}, for a stochastic training algorithm $\mathcal A$ we write $h \sim \ha$ to denote sampling a hypothesis from the distribution induced by the algorithm, \ie by running the algorithm and collecting the hypothesis $h: \mathcal X \to \mathcal Y$ computed by the trained network. We write $\err_{x,y}(h) = 1\{h(x) \neq y\}$ to denote the event that $h$ makes an error on the example $(x, y)$, so that $\E_{h \sim \ha}[\err_{x,y}(h)]$ is the proportion of runs of training which make an error on $(x, y)$. We additionally write $\err(h) = \E_{(x, y) \sim \mathcal D}[\err_{x,y}(h)]$ to denote the distribution-wise error rate, so that the distribution-wise variance is $\Var_{h \sim \ha}(\err(h))$. For a test-set $S = ((x_1, y_1), \dots, (x_n, y_n))$ we define the test-set error as usual as $\err_S(h) = \frac1n \sum_{i=1}^n \err_{x_i,y_i}(h)$. Finally, we write $\err(\mathcal A) = \E_{h \sim \ha}[\err(h)]$ to denote the mean distribution-wise error of hypotheses produced by the training algorithm.
We often refer interchangeably to the variance of the accuracy and of the error rate, as the two quantities always have the same variance.

\textbf{Main experimental setup.} For our main experiments (\Secref{sec:variance}) we train ResNets on CIFAR-10.
We study four different training durations: 0, 4, 16, and 64 epochs. The 0-epoch case corresponds to evaluating the network at initialization; on average this has random chance-level accuracy, but some random initializations reach as high as 14\% and as low as 6\% accuracy.
A complete description of each training configuration is provided in \Secref{app:hyperparams}.
We execute each configuration 60,000 times and collect the resulting test-set predictions, yielding the accuracy distributions shown in \Figref{fig:distributions}. These 240,000 collected sets of test-set predictions form our main object of study.

\newpage

\section{The statistical structure of neural network errors}
\label{sec:variance}
\vspace{-1mm}

\subsection{Do lucky random seeds generalize?}
\label{sec:do_lucky}
\begin{figure}
    \centering
    \includegraphics[width=1.0\textwidth]{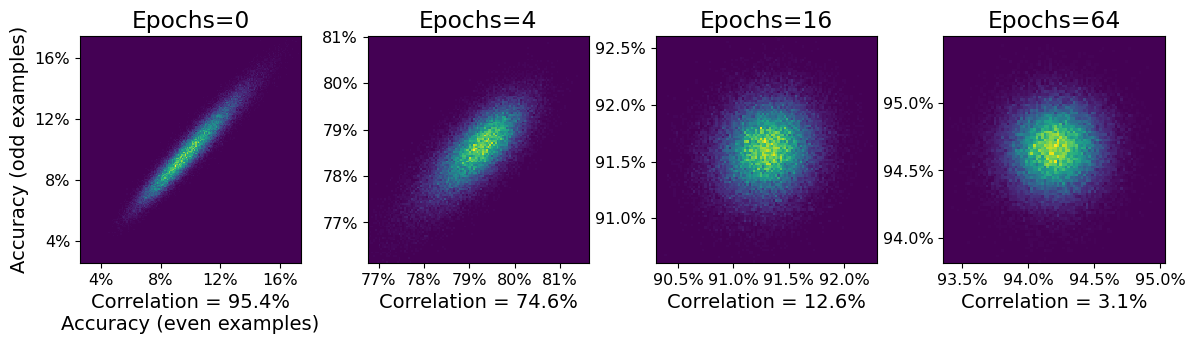}
    % \vspace{-0.5cm}
    \caption{\small \textbf{Error rates on disjoint splits of test data become decorrelated when training to convergence.} We evaluate a large number of independently trained networks on two splits of the CIFAR-10 test-set. When under-training there is substantial correlation, so that a ``lucky'' run which over-performs on the first split is also likely to achieve higher-than-average accuracy on the second. As we increase the training duration, the two error rates decorrelate from each other.}
    \label{fig:split}
    \vspace{-0.2cm}
\end{figure}

In \Figref{fig:distributions} we observed that our standard CIFAR-10 training configuration has significant variation between runs.
Even when training for a long duration, we found pairs of random seeds which produce trained networks whose test-set accuracy differs by more than 1\%.
In this section, we argue that this variance is merely a form of finite-sample noise caused by the limited size of the test-set, and does not imply almost any genuine fluctuation in the quality of the trained network.

Suppose we view the random seed as a training hyperparameter.
Then we have observed that it can be effectively ``tuned'' to obtain improved performance on the test-set --
on average, our training configuration attains an accuracy of $94.42\%$, but we can find random seeds which reach above $95\%$, which is more than a 10\% reduction in the number of errors. However, this improvement on the test-set alone is not enough to conclude that the random seed genuinely affects model quality. What remains to be seen is whether this performance improvement can generalize to unseen data, or if we are just effectively over-fitting the random seed to the observed test-set.

To find out, we perform the following experiment. First, we split the CIFAR-10 test-set into two halves of 5,000 examples each. CIFAR-10 is already shuffled, so for convenience we simply use the odd and even-indexed examples as the two halves. We view the first half as the hyperparameter-validation split and second as the held-out test split. Next, we execute many independent runs of training, with identical configurations other than the varying random seed. We measure the performance of each trained network on both splits of data. If lucky random seeds do generalize, then runs which perform well on the first split should also perform better than average on the second split. 

To additionally determine the effect of training duration, we repeat this experiment for trainings of 0, 4, 16, and 64 epochs, using 60,000 independently trained networks for each duration.
We view the results in \Figref{fig:split}.
For short trainings, the two splits are indeed highly correlated, such that runs which perform well on the first split also tend to do well on the second. But when training for longer, this correlation nearly disappears. For example, when training for 64 epochs, our highest-performing network on the first split does not even perform better than average on the second split. And on average, the top $1/4$ of runs with respect to the first split perform only 0.02\% better than average on the second split.

This result has the following practical implication. Suppose we want to obtain a good CIFAR-10 model. Noticing significant variation between runs (\Figref{fig:distributions}), we might be tempted to re-run training many times, in order to obtain networks with better test-set performance. However, according to \Figref{fig:split}, this would be useless, because improvements on the test-set due to re-training will have near-zero correlation with improvements on unseen data. These networks would be ``better'' only in the sense of attaining higher test-set accuracy, but not in the sense of being more accurate on unseen data from the same distribution.

% \vspace{-1mm}
\subsection{Errors are approximately independent}
\label{sec:hypothesis}
% \vspace{-1mm}

\begin{figure}
    \centering
    \includegraphics[width=.97\textwidth]{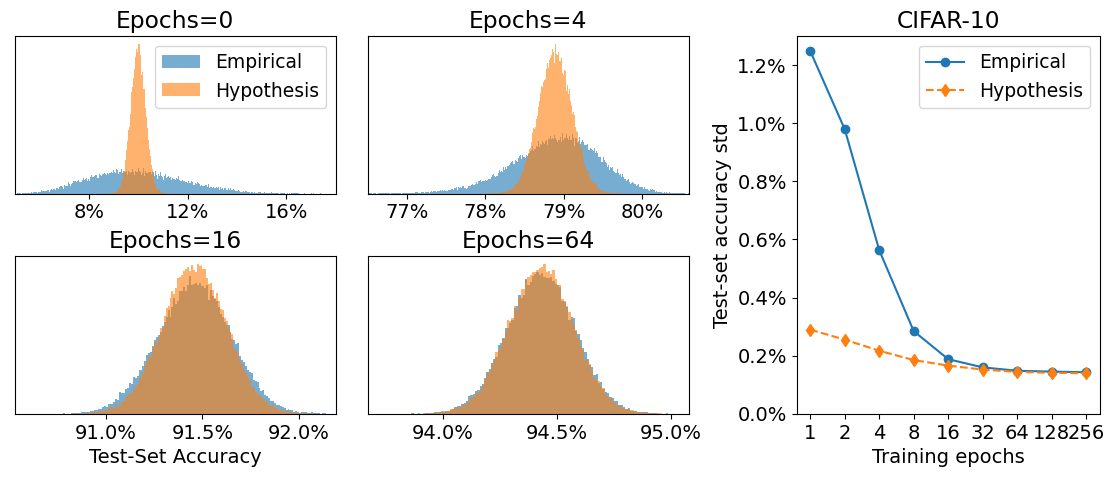}
    \vspace{-1.5mm}
    \caption{\small \textbf{Independent errors explain variance when training to convergence.} (Left:) We compare the empirical distribution of test-set accuracy with that generated by simulating an equal number of samples assuming the hypothesis of independent errors. The hypothesis is wrong for short trainings, but becomes a close fit as training progresses. (Right:) The hypothesis accurately predicts variance when training to convergence.}
    \label{fig:hypothesis}
    \vspace{-3mm}
\end{figure}

In the previous section we showed that when training to convergence, disjoint splits of test data become nearly decorrelated, in the sense that networks which randomly perform well on one split do not perform better than average on the other.
We now test the hypothesis that this phenomenon
\begin{wrapfigure}{r}{0.45\textwidth}
\begin{center}
    \vspace{-4mm}
    \includegraphics[width=0.22\textwidth]{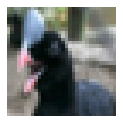}
    \includegraphics[width=0.22\textwidth]{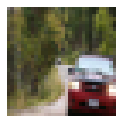}
    \vspace{-5mm}
    \caption{\small \textbf{A pair with independent errors.} (Left) is image 776 of the CIFAR-10 test-set. Out of 60,000 independent runs of 64-epoch training, 21,736 networks (36.2\%) correctly predict this example. (Right) is image 796, which is correctly predicted by 36,392 networks (60.7\%). The number of networks which predict both correctly at the same time is 13,103 (21.83\%), which has a statistically insignificant difference to the quantity $0.362 \cdot 0.607 = 21.97\%$, which is the predicted value if their errors are independent.}
    % \vspace{-1.2cm}
    \vspace{-5mm}
    % \vspace{-8mm}
    \label{fig:example_pair}
\end{center}
\end{wrapfigure}
also extends to individual examples, such that the event that the network makes an error on a given test example becomes independent of its other errors as we train to convergence.
\Figref{fig:example_pair} shows an example.

\begin{definition}
\label{def:independent}
The training algorithm $\mathcal A$ makes\\ \textbf{independent errors} on a test-set $S$ if for every pair of examples $(x_i, y_i)$ and $(x_j, y_j)$ in $S$,
\begin{equation}
\oCov_{h \sim \ha}(\err_{x_i,y_i}(h), \err_{x_j,y_j}(h)) = 0
\end{equation}
\end{definition}
% \vspace{-2mm}
For classification problems, each example's error event $\err_{x_i,y_i}(h) = 1\{h(x_i) \neq y_i\}$ is a Bernoulli variable over the training stochasticity. If the training algorithm makes independent errors, then these error variables form a series of independent biased coin flips, with the test-set error rate distributed as their average.
To find out whether this hypothetical distribution matches empirical reality, we first collect the example-wise mean error rates $\varepsilon_i := \E_h[\err_{x_i,y_i}(h)]$ for $i \in \{1, \dots, n\}$, and then sample from the distribution by taking the average of $n$ coin flips which are biased by those rates.
The exact variance of the resulting random variable is given by the formula $\frac{1}{n^2}\sum_{i=1}^n \varepsilon_i(1-\varepsilon_i)$.

In \Figref{fig:hypothesis} we compare this simulation of the hypothesized distribution to the empirical distribution generated by repeated runs of training. We find that for short trainings of 0-16 epochs, the hypothesis is wrong, as the empirical distribution has extra variance which the hypothesis does not account for. But for the full 64-epoch training, the hypothesis becomes a close fit to reality. It predicts a standard deviation of 0.145\% and the empirical value is similar at 0.149\%.

As further confirmation of the hypothesis, we find that there exist only five pairs of examples in the CIFAR-10 test-set which deviate from having independent errors by more than 2\%. We show all five pairs in \Figref{fig:anompair}. We additionally show in \Figref{fig:binomial} that the hypothesis distribution compares favorably to that generated by the binomial assumption.
In the next section we explore how the small difference between the independent errors hypothesis and reality can be used to estimate the variance in accuracy with respect to the underlying test-distribution.

\subsection{Distribution-wise variance is small}
\label{sec:estimating_distrib}
In \Secref{sec:do_lucky} we showed that accuracy is decorrelated between disjoint splits of test-data, and argued that this implies there is little genuine variation in model quality between runs of training.
In this section we formalize our notion of model quality as accuracy on the underlying test-distribution, and show that the variance of this quantity between runs of training turns out to be very small.

Neural networks are typically evaluated by their performance on a test-set. However, what ultimately determines the expected performance of a neural network on new batches of unseen data is not the test-set error rate, but rather the error rate on the underlying distribution from which the test-set was sampled. We call this the distribution-wise error rate $\err(h) = \E_{(x, y) \sim \mathcal D}[1\{h(x) \neq y\}]$.

\begin{wrapfigure}{r}{0.45\textwidth}
\begin{center}
\vspace{-5mm}
    \includegraphics[width=0.44\textwidth]{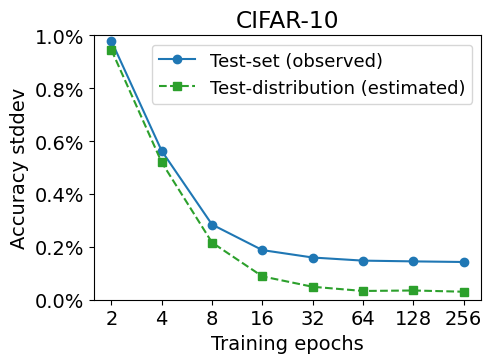}
    \caption{\small \textbf{Test-set variance overestimates distribution-wise variance.} We use \Eqref{eq:estimate_distributionwise} to estimate the distribution-wise variance $\Var_{h \sim \ha}(\err(h))$. It becomes 20$\times$ smaller than the test-set variance when training to convergence.}
    \label{fig:estimate_infinite}
\end{center}
\vspace{-5mm}
\end{wrapfigure}

Estimating the \textit{mean} of distribution-wise error across training stochasticity is relatively easy, because the mean test-set accuracy is an unbiased estimator as $\E_{S \sim \mathcal D^n}[\E_h[\err_S(h)]] = \E_h[\err(h)]$.
Estimating its \textit{variance} $\Var_h(\err(h))$ is more challenging, because the variance in test-set accuracy is potentially an overestimate (proof in \Secref{app:proof1}).
\begin{theorem}
\label{thm:thm1}
In expectation, the variance in test-set accuracy overestimates the variance in true error.
\begin{equation}
\oE_{S \sim \mathcal{D}^n}\left[\oVar_{h \sim \ha}(\err_S(h))\right] \geq \oVar_{h \sim \ha}(\err(h))
\end{equation}
\end{theorem}

We note that this $\geq$ becomes a strict $>$ given very mild additional assumptions. To obtain an unbiased estimate of $\oVar_h(\err(h))$, we recall the results of \Secref{sec:hypothesis}.
When training to convergence, test-set accuracy follows a distribution which can be approximately recovered by assuming that errors are independent (Definition~\ref{def:independent}, \Figref{fig:hypothesis}). The distribution-wise variance in this case should be essentially zero, assuming the distribution is not concentrated in a small number of discrete examples. On the other hand, for shorter trainings we observed substantial test-set variance in excess of that predicted by Definition~\ref{def:independent}, \ie $\oVar_{h \sim \ha}(\err_S(h)) > \frac{1}{n^2}\sum_{i=1}^n \oVar_h(\err_{x_i,y_i}(h))$. For example, the hypothesis predicts that our 4-epoch configuration should have a standard deviation of 0.22\%, but the empirical value was much larger at 0.56\%. This suggests that these shorter trainings may have significant distribution-wise variance between runs.

We provide the following theorem which makes this intuitive connection rigorous (proof in \Secref{app:proof2}). It turns out that a slight rescaling of the excess in empirical variance over that predicted by Definition~\ref{def:independent} forms an unbiased estimator for the variance of the distribution-wise error rate.
\begin{theorem}
\label{thm:thm2}
The following quantity is an unbiased estimator for $\oVar_{h \sim \ha}(\err(h))$.
\begin{equation}\label{eq:estimate_distributionwise}
\hat{\sigma}_S^2 = \frac{n}{n-1}\left(\oVar_{h \sim \ha}(\err_S(h)) \,-\, \frac{1}{n^2} \sum_{i=1}^n \oVar_{h \sim \ha}(\err_{x_i, y_i}(h))\right)
\end{equation}
\end{theorem}
In \Figref{fig:estimate_infinite} we compare this quantity to the test-set variance $\oVar_{h \sim \ha}(\err_S(h))$ across a range of training durations.
When training for 4 epochs, the standard deviation of distribution-wise error is estimated at $\sqrt{\hat\sigma_S^2} = 0.52\%$, indicating that there are indeed significant differences in quality between neural networks trained for this duration. In contrast, when training for the full duration of 64 epochs, we estimate that \textit{the distribution-wise error rate has a standard deviation between runs of only 0.033\%}. In \Secref{sec:imagenet} we obtain a similarly small estimate for ImageNet trainings. This is 248$\times$ less variance than the 4-epoch configuration has, and 20$\times$ less variance than the 64-epoch configuration naively has on the test-set.
This result indicates that when training to convergence, there is very little variation in model quality (\ie expected performance with respect to new batches of data from the test-distribution) between runs of training.

Having confirmed that the distribution-wise variance is small, it still remains to explain why there is high variance on the finite test-set in the first place. We investigate this question in the next section.

\subsection{Variation can be predicted from class-wise calibration}
\label{sec:ece_bound}

In this section we prove that variance in test-set accuracy between runs of training is an inevitable consequence of the fact that ensembles of trained networks approximately satisfy the class-wise calibration property of \citet{jiang2021assessing}. Our analysis yields a simple formula which accurately estimates variance for binary classification problems.

\begin{figure}
    \centering
    \includegraphics[height=3.5cm]{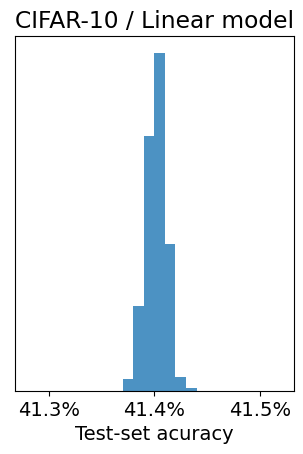}
    \includegraphics[height=3.5cm]{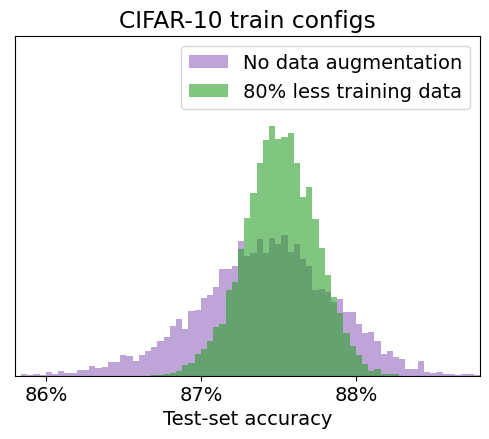}
    \includegraphics[height=3.5cm]{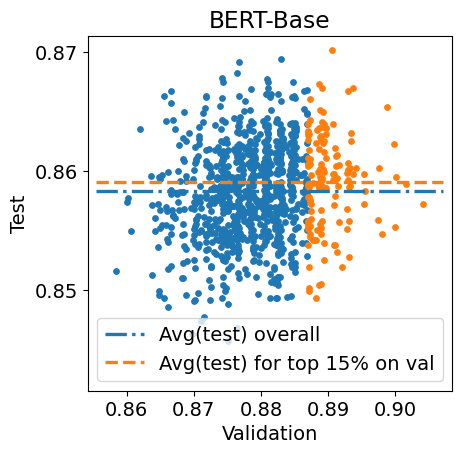}
    \includegraphics[height=3.5cm]{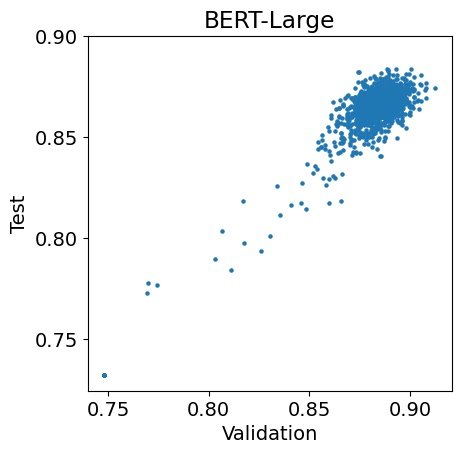}
    \caption{\small (Far left:) A regularized linear model has very little variance between runs of training. (Center left:) Removing either data augmentation or 80\% of training data reduces the mean accuracy to a similar level, but the former produces far more variance than the latter. (Right two:) When finetuning BERT$_\text{BASE}$ on MRPC, performance on the validation and test sets are not strongly correlated across repeated runs. On the other hand, BERT$_\text{LARGE}$ has significant correlated variance, indicating genuine distribution-wise variance.\vspace{-0.3cm}}
    \label{fig:assorted}
\end{figure}

Classical machine learning algorithms based on convex loss functions can have neither test-set-wise nor test-distribution-wise variance, because training eventually converges to the single global optimum which always makes the same set of predictions. For example, in \Figref{fig:assorted} (left) we find that repeatedly training a regularized linear model on CIFAR-10 leads to a standard deviation of below 0.01\%. On the other hand, neural networks can have many optima~\citep{auer1995exponentially,choromanska2015loss}, so that every run of training can potentially lead to a different solution with different behavior on the test-distribution. Despite this, in the previous section we showed that neural network trainings in fact have little variance in their overall distribution-wise performance. What then explains the property of neural network trainings that they have high variance on their test-sets?

We argue that the following property of neural network trainings, which \citet{jiang2021assessing} demonstrate approximately holds in practice, is connected to their variable performance on test-sets.
\begin{definition}
\label{def:calibration}
The stochastic training algorithm $\mathcal A$ satisfies \textbf{class-wise calibration}~\citep{jiang2021assessing} if for every class $c \in \mathcal Y$ and confidence level $q \in [0, 1]$,
\begin{equation}
P_{(x, y) \sim \mathcal D}\big(y = c \mid P_{h \sim \ha}(h(x) = c) = q\big) = q.
\end{equation}
\end{definition}

As an explanatory example, if we let $S' \subset S$ be the subset of test images which are classified by 30-40\% of independently trained neural networks as ``cat,'' then 30-40\% of $S'$ really will be cats.

We prove (\Secref{app:proof3}) the following theorem connecting class-wise calibration to variance.
% We connect class-wise calibration to test-set variance with the following theorem (proof in \Secref{app:proof3}).

\begin{theorem}
\label{thm:thm3}
Let $\mathcal A$ be a stochastic training algorithm for binary classification. If it is class-wise calibrated, then its expected variance on a test-set of size $n$ is equal to
\begin{equation}
\oE_{S \sim \mathcal{D}^n}\left[\oVar_{h \sim \ha}(\err_S(h))\right] = \frac{\err(\mathcal A)}{2n} \,\,+\,\, (1 \!-\! \tfrac1n) \cdot \oVar_{h \sim \ha}(\err(h))
\end{equation}
\end{theorem}

\begin{wrapfigure}{r}{0.45\textwidth}
\vspace{-4mm}
\begin{center}
    \includegraphics[width=0.44\textwidth]{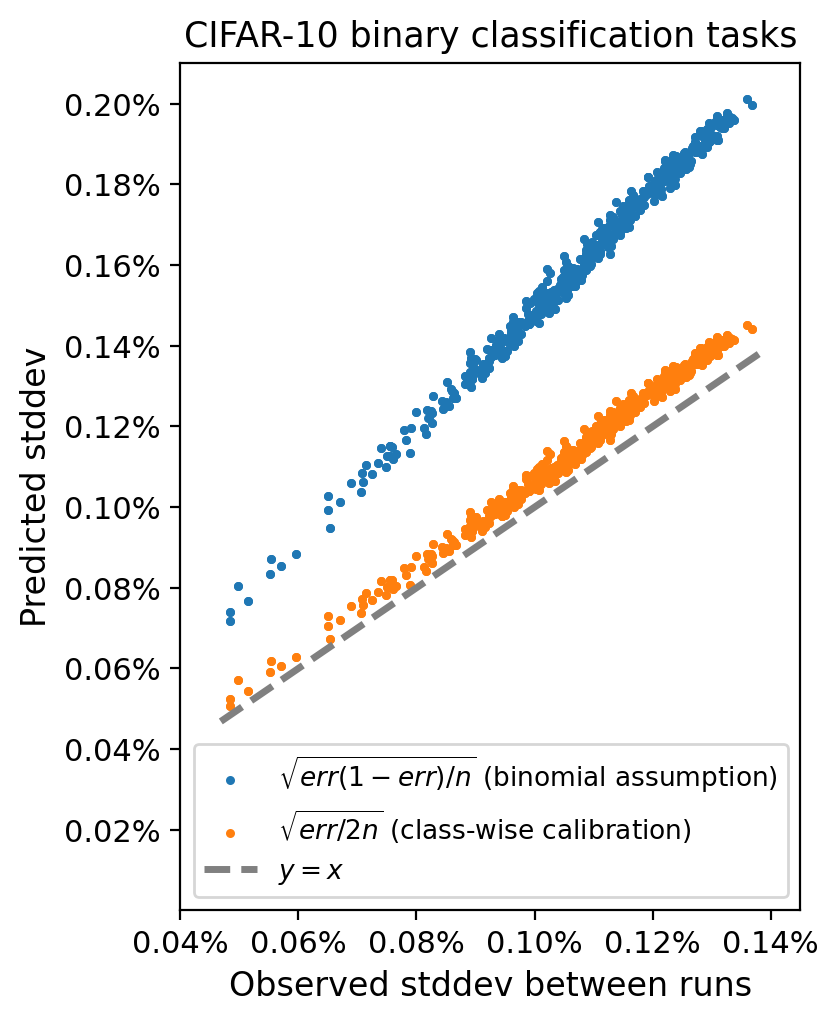}
    % \vspace{-4mm}
    \caption{\small \textbf{Predicting variance.} Across hundreds of tasks, \Eqref{eq:binary} accurately predicts the standard deviation of test-set error. In contrast, the binomial assumption is inaccurate.}
    \label{fig:std_formulas}
\end{center}
\vspace{-2mm}
\end{wrapfigure}
We showed in \Secref{sec:estimating_distrib} that $\Var_h(\err(h))$ is small in practice. Therefore, this theorem practically reduces to the following simple formula:
\begin{equation}
\label{eq:binary}
\E_S[\Var_h(\err_S(h)] \approx \err(\mathcal A)/2n
\end{equation}
In \Figref{fig:std_formulas} we use this formula to predict test-set variance across 511 different binary classification tasks generated by assigning each CIFAR-10 class to be either positive or negative. We use each task's test-set error rate as a cheap approximation to $\err(\mathcal A)$. The resulting predictions are a close fit with empirical reality, with $R^2 = 0.996$ across the collection of tasks.
For example, compared to the commonly used binomial assumption~\citep{dietterich1998approximate,raschka2018model}, the variances predicted by \Eqref{eq:binary} are $70\times$ more accurate in terms of their mean squared distance to the empirical values.

We additionally prove (\Secref{app:proof4}) a lower bound for general $k$-way classification.
\begin{theorem}
\label{thm:thm4}
Given a training algorithm $\mathcal A$ for $k$-way classification, if it is class-wise calibrated, then its expected variance on a test-set of size $n$ is at least
\begin{equation}
\oE_{S \sim \mathcal{D}^n}\left[\oVar_{h \sim \ha}(\err_S(h))\right] \geq \dfrac{\err(\mathcal A)}{nk}
\end{equation}
\end{theorem}
Together, these theorems show that the variance of neural network trainings on finite test-sets is a predictable consequence of their class-wise calibration.

\section{Additional experiments}
\label{sec:experiments}

In this section we conduct preliminary investigations regarding the effect of data augmentation, learning rate, finetuning instability, and distribution-shift on variance. We additionally include a replication study of \citet{summers2021nondeterminism} in \Secref{sec:factors}, which confirms their findings that variance is caused by extreme sensitivity to initial conditions rather than any particular stochastic factor like network initialization or data ordering.
% Finally, we present a novel kernel function between datapoints in Appendix~\ref{sec:similarity}.

\vspace{-1mm}
\subsection{The effect of finetuning instability}
\vspace{-1mm}
\label{sec:bert_finetune}
In this section we study BERT~\citep{devlin2018bert} finetuning, a setting where previous works have reported significant variance between runs~\citep{devlin2018bert,dodge2020fine,mosbach2020stability}. Our contribution is to use the tools developed in \Secref{sec:variance} to clearly differentiate the behavior of BERT$_\text{LARGE}$ from BERT$_\text{BASE}$.

For our experiment, we finetune pretrained checkpoints of both models 1,000 times each on the MRPC~\citep{dolan2005automatically} task.
In \Figref{fig:assorted} (right) we show that for BERT$_\text{BASE}$, the validation and test splits of MRPC are close to decorrelated in terms of the finetuned model performance, similarly to \Secref{sec:do_lucky}. The top 15\% of seeds in terms of validation-set performance achieve only 0.09\% higher performance than average on the test-set, whereas for BERT$_\text{LARGE}$ the correlation is higher.
The test-set error rate of BERT$_\text{BASE}$ has a standard deviation of 0.80\% between runs of finetuning, and BERT$_\text{LARGE}$ has a stddev of 2.24\%.
This is already a significant gap of almost $8\times$ more test-set variance for BERT$_\text{LARGE}$; using \Eqref{eq:estimate_distributionwise} increases the gap further. In particular, we estimate the distribution-wise standard deviation at 0.21\% for BERT$_\text{BASE}$, and 2.08\% for BERT$_\text{LARGE}$, amounting to $100\times$ more distribution-wise variance for BERT$_\text{LARGE}$.

\vspace{-1mm}
\subsection{The effect of data augmentation}
\vspace{-1mm}
\label{sec:data_augs}
In this section we investigate the effect of data augmentation on variance. In \Figref{fig:assorted} (center left) we compare two modifications of our CIFAR-10 training: first, removing a fixed 80\% of training data, and second, removing data augmentation.
While both modifications yield a similar mean accuracy of 87.5\%, removing augmentations results in 3.5$\times$ more variance between runs. Furthermore, the large-ensemble accuracy of the networks trained without augmentation is higher, reaching 91.2\%, compared to the reduced-data ensemble, which reaches only 89.8\%. We conclude that data augmentation reduces variance, although the mechanism by which this happens is not yet known.

\subsection{The effect of learning rate}
\label{sec:learning_rate}
\begin{figure}
    \centering
    \includegraphics[height=4.5cm]{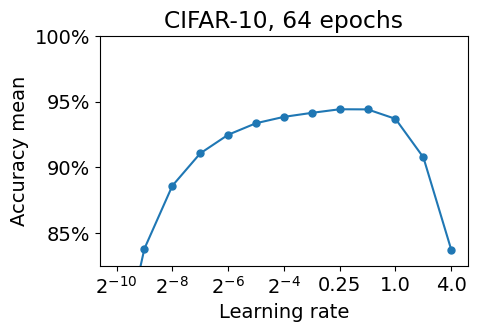}
    \hspace{3mm}
    \includegraphics[height=4.5cm]{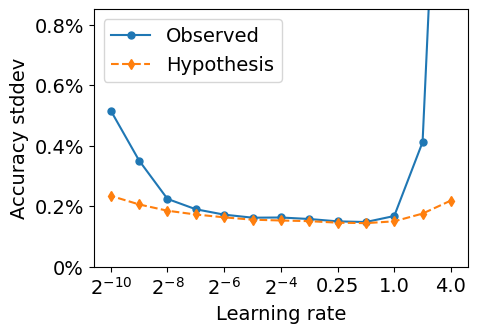}
    \vspace{-2mm}
    \caption{\small \textbf{Accuracy is maximized by the largest learning rate without excess variance.} Across learning rates, we compare the observed stddev of test-set accuracy to that predicted by the independent errors framework. The best learning rate is apparently the largest one which does not induce significant excess variance.}
    \vspace{-3mm}
    \label{fig:learning_rate}
\end{figure}

In this section we investigate the relationship between learning rate and variance. Our experiment is to execute 1,000 64-epoch CIFAR-10 trainings for each binary-power learning rate between $2^{-10}$ and $2^2$. For each setting, we measure the mean and variance of test-set accuracy.
We observe that the learning rate 0.5 yields both the highest mean and the lowest variance. Raising it to 1.0 causes the standard deviation of test-set accuracy to increase from 0.148\% to 0.168\%. This may seem insignificant, but \Eqref{eq:estimate_distributionwise} estimates that it implies a significant $5\times$ increase in distribution-wise variance.
We therefore conjecture that, as a general property of neural network trainings, the optimal learning rate is the largest one which does not induce significant distribution-wise variance.

\vspace{-1mm}
\subsection{The effect of distribution shift}
\vspace{-1mm}
\label{sec:shift}
In this section we summarize our results on distribution shift, which are fully described in \Secref{sec:imagenet}. Our experimental setup is to train 1,000 ResNet-18s on ImageNet with identical hyperparameters, and then use \Eqref{eq:estimate_distributionwise} to estimate their distribution-wise variances on various test-sets: namely, the main ImageNet validation set, ImageNet-V2~\citep{recht2019imagenet}, and three distribution-shifted sets. We find that the main validation set has a distribution-wise standard deviation of 0.034\%, which is quite similar to the CIFAR-10 case. ImageNet-V2, which is intended to be a sample from the same underlying distribution as ImageNet, has a standard deviation of 0.071\%. Each of the three distribution-shifted sets, on the other hand, have at least six times more variance than that.
For example, ImageNet-Sketch~\citep{wang2019learning} has a distribution-wise standard deviation of 0.257\%, meaning that on average every 50 runs will contain a pair of networks whose performance on the ImageNet-Sketch distribution differs by over 1\%, assuming a roughly Gaussian accuracy distribution.
Overall, distribution-wise variance is high for precisely those test-distributions which are significantly shifted relative to the training distribution.
Why this happens is unknown.

\vspace{-1mm}
\section{Discussion}
\vspace{-1mm}

A central focus of this paper is the distinction between a model's observed error rate on a test-set, and its true error rate on the underlying test-distribution from which that test-set was sampled.
The mean of the former, over repeated runs of training, provides an unbiased estimate of the mean of the latter. But the variance of the former does not in general provide an unbiased estimate for the variance of the latter, meaning that we cannot tell how unstable a training really is just from measuring its variance on a test-set. To recover an unbiased estimate, we derive a new formula (\Eqref{eq:estimate_distributionwise}). Using this formula we discover that for standard trainings, the variance between runs of training in terms of distribution-wise error rate is typically very small, with a standard deviation of only around 0.03\% for both CIFAR-10 and ImageNet trainings.

Our understanding of variance is further simplified by the \textit{independent errors} framework. It turns out that over repeated runs of standard CIFAR-10 training,
the event that the trained network makes an error on one particular example has almost no effect on its chances of making other errors.
Overall, our takeaway is that for standard trainings, \textit{even though some random seeds lead to substantially higher or lower performance on the test-set due to independent errors, all seeds have nearly equal performance on the underlying test-distribution}.

However, we found two exceptions to this takeaway. The first is trainings which have pathological instability, such as BERT$_{\text{LARGE}}$ finetuning where the test-set accuracy can vary by more than 15\%~(\Secref{sec:bert_finetune}), and distribution-wise variance is also high. The second more interesting exception is trainings whose test distributions are shifted relative to their training distributions~(\Secref{sec:shift}, \Secref{sec:imagenet}).
% We observed distribution-wise variance to be many times higher for this case~(\Secref{sec:shift}, \Secref{sec:imagenet}).
Understanding why variance appears alongside distribution shift is an intriguing task whose solution we look forward to in future work.

\subsection*{Acknowledgements}
We are grateful to Behnam Neyshabur for his guidance on a preliminary version of this work. We thank
Ehsan Amid, Luke Johnston, and Ryan Weber each for their insightful comments on the draft.

\bibliography{iclr2024_conference}
\bibliographystyle{iclr2024_conference}

\newpage
\appendix
\section{Training details}
\label{app:hyperparams}

\subsection{CIFAR-10}

For our main experiments (\Secref{sec:variance}) we train thousands of ResNet-9s on CIFAR-10. Our network architecture is the same as was used by \citet{ilyas2022datamodels}, namely, a 9-layer ResNet descended from \citet{paged2019resnet}. Our 0-epoch configuration corresponds to a randomly initialized network. Our 4, 16, and 64-epoch configurations all train using SGD with learning rate 0.5, momentum 0.9, and weight decay 5e-4, with the learning rate linearly ramped down to zero by the end of training. We train using random flipping, 2-pixel translation, and 12-pixel Cutout~\citep{devries2017improved} data augmentations. We use batch size 500 and load data using the FFCV~\citep{leclerc2022ffcv} library. Our training script is made available at \url{https://github.com/KellerJordan/ffcv-cifar/blob/master/train.py}. The 64-epoch configuration attains an average accuracy of 94.42\% without the use of test-time augmentation. For each of the four durations we execute training 60,000 times, generating 240,000 sets of test-set predictions, which form our object of study for \Secref{sec:variance}.

\subsection{ImageNet}
\label{app:hparams_in}
For our ImageNet experiments (\Secref{sec:shift}, \Secref{sec:imagenet}) we train 1,000 ResNet-18s on ImageNet. We use standard random flip and random resized crop data augmentations. We train at resolution 192 for 100 epochs with batch size 1024, using SGD-momentum with learning rate 0.5, momentum 0.9, and weight decay 5e-5. We linearly ramp the learning rate up from 5e-5 to 0.5 by epoch 2, and then down to zero by the end of training. We evaluate at resolution 256 with crop ratio 0.875. We use the FFCV dataloader here as well, and base our training on \url{https://github.com/libffcv/ffcv-imagenet}.

\subsection{BERT finetuning}
For our BERT finetuning experiments (\Secref{sec:bert_finetune}) we finetune BERT$_\text{BASE}$ and BERT$_\text{LARGE}$ on the MRPC~\citep{dolan2005automatically} binary classification dataset. MRPC contains 3,668 training examples, 407 validation examples, and 1,725 test examples. We train for 3 epochs at batch size 16, using Adam~\citep{kingma2014adam} with default hyperparameters other than the learning rate, which is linearly ramped from a maximum of 2e-5 down to zero by the end of training.

\newpage
\section{Proofs}
\label{app:proofs}

\begin{lemma}
\label{lem:lem1}
The variance of error is equal to the expected covariance between pairs of examples.
\[
\oVar_{h \sim \ha}(\err(h)) = \oE_{(x_1, y_1), (x_2, y_2) \sim \mathcal{D}^2}\left[\oCov_{h \sim \ha}\big(\err_{x_1, y_1}(h), \err_{x_2, y_2}(h)\big)\right]
\]
\end{lemma}
\begin{proof}
\begin{align*}
\oVar_{h \sim \ha}(\err(h)) &= \oE_{h \sim \ha}\big[\left(\err(h) - \E_{h'}[\err(h')]\right)^2\big] \\
&= \oE_h\left[\left(\oE_{x, y}[\err_{x, y}(h) - \oE_{h'}[\err_{x, y}(h')]]\right)^2\right] \\
&= \oE_h\left[\oE_{x_1, y_1}[\err_{x_1, y_1}(h) - \oE_{h'}[\err_{x_1, y_1}(h')]] \cdot \oE_{x_2, y_2}[\err_{x_2, y_2}(h) - \oE_{h'}[\err_{x_2, y_2}(h')]]\right] \\
&= \oE_h\oE_{x_1, y_1}\oE_{x_2, y_2}\left[(\err_{x_1, y_1}(h) - \oE_{h'}[\err_{x_1, y_1}(h')])(\err_{x_2, y_2}(h) - \oE_{h'}[\err_{x_2, y_2}(h')])\right] \\
&= \oE_{x_1, y_1}\oE_{x_2, y_2}\oE_h\left[(\err_{x_1, y_1}(h) - \oE_{h'}[\err_{x_1, y_1}(h')])(\err_{x_2, y_2}(h) - \oE_{h'}[\err_{x_2, y_2}(h')])\right] \\
&= \oE_{(x_1, y_1), (x_2, y_2) \sim \mathcal{D}^2}\left[\oCov_{h \sim \ha}\left(\err_{x_1, y_1}(h), \err_{x_2, y_2}(h)\right)\right].
\end{align*}
\end{proof}

% \newpage
\begin{lemma}
\label{lem:lem2}
For an IID test-set $S = ((x_1, y_1), \dots, (x_n, y_n))$, the expected variance in the test error rate can be decomposed into a mixture of distribution-wise and example-wise variances.
\[ \oE_{S \sim \mathcal{D}^n}\left[\oVar_{h \sim \ha}(\err_S(h))\right] = (1 - 1/n)\cdot\oVar_{h \sim \ha}(\err(h)) \,+\, (1/n)\cdot\oE_{(x, y) \sim \mathcal D}\left[\oVar_{h \sim \ha}(\err_{x, y}(h)) \right] \]
\end{lemma}
\begin{proof}
\begin{align*}
&\oE_{S \sim \mathcal{D}^n}\left[\oVar_{h \sim \ha}(\err_S(h))\right] = \oE_S\left[\oVar_h\left(\frac{1}{n}\sum_{i=1}^n \err_{x_i, y_i}(h)\right) \right] \\
&= \oE_S\left[\frac{1}{n^2}\sum_{i=1}^n\sum_{j=1}^n \oCov_h(\err_{x_i, y_i}(h), \err_{x_j, y_j}(h)) \right] \\
&= \frac{1}{n^2}\sum_{i=1}^n\sum_{j=1}^n \oE_S\left[\oCov_h(\err_{x_i, y_i}(h), \err_{x_j, y_j}(h))\right] \\
&= \frac{1}{n^2}\sum_{i=1}^n \oE_S\left[ \oVar_h(\err_{x_i, y_i}(h)) \right] + \frac{1}{n^2}\sum_{i=1}^n\sum_{j \neq i} \oE_S\left[\oCov_h(\err_{x_i, y_i}(h), \err_{x_j, y_j}(h)) \right] \\
&= \frac{1}{n^2}\sum_{i=1}^n \oE_{x_i,y_i}\left[\oVar_h(\err_{x_i, y_i}(h))\right] + \frac{1}{n^2}\sum_{i=1}^n\sum_{j \neq i} \oE_{(x_i,y_i),(x_j,y_j)}\left[\oCov_h(\err_{x_i, y_i}(h), \err_{x_j, y_j}(h))\right] \\
&= \frac{n}{n^2}\oE_{x, y}\left[\oVar_h(\err_{x, y}(h)) \right] + \frac{n(n-1)}{n^2}\oE_{(x_1, y_1), (x_2, y_2)}\left[\oCov_h(\err_{x_1, y_1}(h), \err_{x_2, y_2}(h)) \right] \\
&= (1/n)\oE_{(x, y) \sim \mathcal D}\left[\oVar_{h \sim \ha}(\err_{x, y}(h)) \right] \,+\, (1 - 1/n)\oE_{(x_1, y_1), (x_2, y_2)}\left[\oCov_h(\err_{x_1, y_1}(h), \err_{x_2, y_2}(h)) \right] \\
% &= \frac{\err(\mathcal A)}{2n} \,+\, (1 - 1/n)\cdot\oE_{(x_1, y_1), (x_2, y_2)}\left[\oCov_h(\err_{x_1, y_1}(h), \err_{x_2, y_2}(h)) \right]
&= \,\,(1/n)\cdot\oE_{(x, y) \sim \mathcal D}\left[\oVar_{h \sim \ha}(\err_{x, y}(h)) \right] \,+\, (1 - 1/n)\cdot\oVar_{h \sim \ha}(\err(h)).
\end{align*}
Where the last step uses Lemma~\ref{lem:lem1}.
\end{proof}

\subsection{Theorem 1}
\label{app:proof1}
\begin{apptheorem}
In expectation, variance in test-set accuracy overestimates variance in true error.
\[ \oE_{S \sim \mathcal{D}^n}\left[\oVar_{h \sim \ha}(\err_S(h))\right] \geq \oVar_{h \sim \ha}(\err(h)) \]
\end{apptheorem}
\begin{proof}
The difference between the two terms is
\begin{align*}
&\oE_{S \sim \mathcal{D}^n}\left[\oVar_{h \sim \ha}(\err_S(h))\right] - \oVar_{h \sim \ha}(\err(h)) \\
&= \frac1n \left(\oE_{(x, y) \sim \mathcal D}\left[\oVar_{h \sim \ha}(\err_{x, y}(h)) \right] - \oVar_{h \sim \ha}(\err(h))\right) \\
&=\frac1n \left(\oE_{x, y}\left[\oVar_h(\err_{x, y}(h)) \right] - \oE_{(x_1, y_1), (x_2, y_2)}\left[\oCov_h(\err_{x_1, y_1}(h), \err_{x_2, y_2}(h)) \right]\right) \\
&=\frac1n \bigg(0.5 \cdot \oE_{x_1, y_1}\bigg[\oVar_h(\err_{x_1, y_1}(h)) \bigg] + 0.5 \cdot \oE_{x_2, y_2}\bigg[\oVar_h(\err_{x_2, y_2}(h)) \bigg] \\
&\quad - \oE_{(x_1, y_1), (x_2, y_2)}\left[\oCov_h(\err_{x_1, y_1}(h), \err_{x_2, y_2}(h)) \right]\bigg) \\
&=\frac{1}{2n} \oE_{(x_1, y_1), (x_2, y_2)}\bigg[\oVar_h(\err_{x_1, y_1}(h)) + \oVar_h(\err_{x_2, y_2}(h)) - 2\oCov_h(\err_{x_1, y_1}(h), \err_{x_2, y_2}(h))\bigg] \\
& \geq \frac{1}{2n} \oE_{(x_1, y_1), (x_2, y_2)}\bigg[\oVar_h(\err_{x_1, y_1}(h)) + \oVar_h(\err_{x_2, y_2}(h)) - 2\sqrt{\oVar_h(\err_{x_1, y_1}(h))\oVar_h(\err_{x_2, y_2}(h))}\bigg] \\
& = \frac{1}{2n} \oE_{(x_1, y_1), (x_2, y_2)}\bigg[\left(\sqrt{\oVar_h(\err_{x_1, y_1}(h))} - \sqrt{\oVar_h(\err_{x_2, y_2}(h))}\right)^2\bigg] \\
& \geq 0.
\end{align*}
Where the first two steps use Lemma~\ref{lem:lem1} and then Lemma~\ref{lem:lem2}.
\end{proof}
Note that this is almost always a strict greater-than, unless the test distribution is a single dirac delta.

\subsection{Theorem 2}
\label{app:proof2}
\begin{apptheorem}
The following quantity is an unbiased estimator for $\oVar_{h \sim \ha}(\err(h))$.
\[
\hat{\sigma}_S^2 = \frac{n}{n-1}\left(\oVar_{h \sim \ha}(\err_S(h)) \,-\, \frac{1}{n^2} \sum_{i=1}^n \oVar_{h \sim \ha}(\err_{x_i, y_i}(h))\right)
\]
\end{apptheorem}
\begin{proof}
\begin{align*}
\oVar_{h \sim \ha}(\err(h)) &= \frac{n}{n-1}\left(\oE_{S \sim \mathcal{D}^n}\left[\oVar_{h \sim \ha}(\err_S(h))\right] \,-\, (1/n)\cdot\oE_{(x, y) \sim \mathcal D}\left[\oVar_{h \sim \ha}(\err_{x, y}(h)) \right]\right) \\
&= \frac{n}{n-1}\left(\oE_{S \sim \mathcal{D}^n}\left[\oVar_{h \sim \ha}(\err_S(h))\right] \,-\, (1/n)\cdot\oE_{S \sim \mathcal D^n}\left[\frac1n \sum_{i=1}^n \oVar_{h \sim \ha}(\err_{x_i, y_i}(h)) \right]\right) \\
&= \oE_{S \sim \mathcal D^n}\left[\frac{n}{n-1}\left(\oVar_{h \sim \ha}(\err_S(h)) \,-\, \frac{1}{n^2} \sum_{i=1}^n \oVar_{h \sim \ha}(\err_{x_i, y_i}(h))\right)\right] \\
\end{align*}
Where the first equality is a rearrangement of Lemma~\ref{lem:lem2}.
\end{proof}

The quantity $\hat{\sigma}_S^2$ is also equal to $\binom{n}{2}^{-1} \sum_{i=1}^n \sum_{j \neq i} \oCov_{h \sim \ha}(\err_{x_i, y_i}(h), \err_{x_j, y_j}(h))$. Comparing this formula to Lemma~\ref{lem:lem1} may help provide intuition for why it is an estimator for the distribution-wise variance. The formulation given in Theorem~\ref{thm:thm2} looks less intuitive, but the benefit is that we only have to calculate $n$ separate variances, rather than $\binom{n}{2}$ covariances.

We note that the proofs of Theorem~\ref{thm:thm1} and Theorem~\ref{thm:thm2} do not assume anything about the error function $\err_{x,y}(h)$\footnote{Other than it being non-pathological enough to allow the interchanges of expectation via Fubini's theorem.}, so, \eg they are also true for regression tasks.

\subsection{Theorem 3}
\label{app:proof3}
\begin{apptheorem}
If $\mathcal A$ is a training algorithm for binary classification which satisfies class-wise calibration (Definition~\ref{def:calibration}), then its expected variance on an IID test-set of size $n$ is
\[
\oE_{S \sim \mathcal{D}^n}\left[\oVar_{h \sim \ha}(\err_S(h))\right] = \frac{\err(\mathcal A)}{2n} \,\,+\,\, (1 - 1/n)\cdot\oVar_{h \sim \ha}(\err(h))
\]
\end{apptheorem}
\begin{proof}
Define the random variable $q(x) = \oE_{h \sim \ha}[1\{h(x) = 1\}]$ to be the proportion of training runs which classify $x$ as positive, with the randomness being over $x \sim \mathcal D$. We first obtain a formula for $\err(\mathcal A)$ in terms of $q$. By the usual laws of conditional expectation we have:
\begin{align*}
\err(\mathcal A) &= \oE_{x,y,h}[\err_{x,y}(h)] = \oE_q[\oE_{x,y,h}[\err_{x,y}(h) \mid q(x) = q]] \\
&= \oE_q[\oE_{x,y,h}[1\{h(x) \neq y\} \mid q]] \\
&= \oE_q[\oE_{x,y,h}[1\{y = 0\}1\{h(x) = 1\} + 1\{y = 1\}1\{h(x) = 0\} \mid q]] \\
&= \oE_q[\oE_{x,y}[\oE_h[1\{y = 0\}1\{h(x) = 1\} + 1\{y = 1\}1\{h(x) = 0\} \mid q] \mid q]] \\
&= \oE_q[\oE_{x,y}[1\{y=0\}\oE_h[1\{h(x) = 1\} \mid q] + 1\{y=1\}\oE_h[1\{h(x) = 0\} \mid q] \mid q]] \\
&= \oE_q[\oE_{x,y}[q \cdot 1\{y = 0\} + (1-q) \cdot 1\{y = 1\} \mid q]] \\
% &= \oE_q[q(1-\oE_{x,y}[1\{y = 1\}\mid q]) + (1-q)\oE_{x,y}[1\{y = 1\}\mid q]] \\
&= \oE_q[q(1-\oE_{x,y}[1\{y = 1\}\mid q]) + (1-q)\oE_{x,y}[1\{y = 1\}\mid q]].
\end{align*}
Using the assumption of class-wise calibration, this formula simplifies to $\oE_q[2q(1-q)]$.
Next we analyze the example-wise variance. We have:
\begin{align*}
\oE_{x,y}[\oVar_h(\err_{x,y}(h))] &= \oE_q[\oE_{x,y}[\oVar_h(1\{h(x) \neq y)\}) \mid q(x) = q]] \\
&= \oE_q[\oE_{x,y}[\oE_h[1\{h(x) \neq y\}](1 - \oE_h[1\{h(x) \neq y\}]) \mid q]] \\
&= \oE_q[\oE_{x,y}[\oE_h[1\{h(x) = 1\}]\oE_h[1\{h(x) = 0\}] \mid q]] \\
&= \oE_q[\oE_{x,y}[q(1-q) \mid q]] \\
&= \oE_q[q(1-q)].
\end{align*}
Where the second equality uses the formula for variance of a Bernoulli variable. The third equality uses the fact that, regardless of whether $y = 0$ or $y = 1$, the product $\oE_h[1\{h(x) \neq y\}](1 - \oE_h[1\{h(x) \neq y\}])$ is equal to $\oE_h[1\{h(x) = 1\}]\oE_h[1\{h(x) = 0\}]$. The fourth equality applies the assumption of class-wise calibration.

Combining the above two results yields $\oE_{x,y}[\oVar_h(\err_{x,y}(h))] = \err(\mathcal A)/2$. Therefore by Lemma~\ref{lem:lem2} we have:
\begin{align*}
\oE_{S \sim \mathcal{D}^n}\left[\oVar_{h \sim \ha}(\err_S(h))\right] &= (1 - 1/n)\cdot\oVar_{h \sim \ha}(\err(h)) \,+\, (1/n)\cdot\oE_{(x, y) \sim \mathcal D}\left[\oVar_{h \sim \ha}(\err_{x, y}(h)) \right] \\
% &= (1 - 1/n)\cdot\oVar_{h \sim \ha}(\err(h)) \,+\, (1/n)\cdot \oE_q[q(1-q)] \\
&= \frac{\err(\mathcal A)}{2n} \,\,\,+\,\,\, (1 - 1/n)\cdot\oVar_{h \sim \ha}(\err(h)).
\end{align*}
\end{proof}

\subsection{Theorem 4}
\label{app:proof4}
\begin{apptheorem}
If $\mathcal A$ is a training algorithm for $k$-way classification which satisfies class-wise calibration (Definition~\ref{def:calibration}), then its expected variance on an IID test-set of size $n$ is at least
\[
\oE_{S \sim \mathcal{D}^n}\left[\oVar_{h \sim \ha}(\err_S(h))\right] \geq \dfrac{\err(\mathcal A)}{nk}
\]
\end{apptheorem}
\begin{proof}
For each class $c \in \{1, \dots, k\}$, define the random variable $q_c(x) = \oE_{h \sim \ha}[1\{h(x) = c\}]$ to be the proportion of runs of training which classify $x$ as $c$. Let $q(x) = (q_1(x), \dots, q_k(x))$ be the vector of these variables. The laws of conditional expectation yield the following expression for the expected error.
\begin{align*}
\err(\mathcal A) &= \oE_{x,y,h}[\err_{x,y}(h)] \\
&= \oE_q[\oE_{x,y,h}[\err_{x,y}(h) \mid q(x) = q]] \\
&= \oE_q[\oE_{x,y,h}[1\{h(x) \neq y\} \mid q]] \\
&= \oE_q\left[\oE_{x,y,h}\left[\sum_{c=1}^k 1\{y = c\}1\{h(x) \neq c\} \mid q\right]\right] \\
&= \oE_q\left[\oE_{x,y}\left[\oE_h\left[\sum_{c=1}^k 1\{y = c\}1\{h(x) \neq c\} \mid q\right] \mid q\right]\right] \\
&= \oE_q\left[\sum_{c=1}^k (1 - q_c)\oE_{x,y}\left[1\{y = c\} \mid q\right]\right] \\
&= \oE_q\left[\sum_{c=1}^k q_c(1 - q_c)\right].
\end{align*}
The last step uses the assumption of class-wise calibration. We next derive a related expression for the example-wise variance.
\begin{align*}
\oE_{x,y}[\oVar_h(\err_{x,y}(h))] &= \oE_q[\oE_{x,y}[\oVar_h(1\{h(x) \neq y)\}) \mid q(x) = q]] \\
&= \oE_q[\oE_{x,y}[\oE_h[1\{h(x) \neq y\} \mid q](1 - \oE_h[1\{h(x) \neq y\} \mid q]) \mid q]] \\
&= \oE_q[\oE_{x,y}[q_y(1 - q_y) \mid q]] \\
&= \oE_q\left[\oE_{x,y}\left[\sum_{c=1}^k 1\{y = c\} q_c(1 - q_c) \mid q\right]\right] \\
&= \oE_q\left[\sum_{c=1}^k \oE_{x,y}[1\{y = c\} \mid q] \cdot q_c(1 - q_c)\right] \\
&= \oE_q\left[\sum_{c=1}^k q_c^2(1 - q_c)\right]
\end{align*}

We now analyze the ratio between $\sum_{c=1}^k q_c^2(1-q_c)$ and $\sum_{c=1}^k q_c(1-q_c)$.

Without loss of generality, let $q_1 \leq q_2 \leq \dots \leq q_k$ be in nondecreasing order. Then we have
\begin{align*}
\sum_{c=1}^k q_c^2(1-q_c) &= k \cdot \left(\frac1k \sum_{c=1}^k q_c \cdot q_c(1-q_c)\right) \\
&\geq k \cdot \left(\frac1k \sum_{c=1}^k q_c\right) \left(\frac1k \sum_{c=1}^k q_c(1-q_c)\right) \\
&= \frac1k \sum_{c=1}^k q_c(1-q_c).
\end{align*}
The inequality step is due to an application of Chevychev's sum inequality~\citep{hardy1934inequalities}, which is possible because the series $q_1(1-q_1), \dots, q_k(1-q_k)$ is nondecreasing, which we prove as follows.

We first recall that $\sum_{c=1}^k q_c = 1$, and that we assumed without loss of generality that $q_1 \leq \dots \leq q_k$. For the first $k-1$ terms, the monotonicity of the mapping $x \mapsto x(1-x)$ on the interval $[0, 1/2]$, combined with the fact that $q_c \leq 1/2$ for $c \in \{1, \dots, k-1\}$, implies $q_1(1-q_1) \leq \dots \leq q_{k-1}(1-q_{k-1})$. It remains to show that $q_{k-1}(1 - q_{k-1}) \leq q_k(1-q_k)$. If $q_k \leq 1/2$, then this is again due to the monotonicity of $x \mapsto x(1-x)$ on $[0, 1/2]$. Otherwise if $q_k \geq 1/2$, then combining $q_{k-1} \leq 1 - q_k$ and $(1 - q_k) \leq 1/2$ yields $q_{k-1}(1 - q_{k-1}) \leq (1 - q_k)(1 - (1 - q_k)) = q_k(1-q_k)$. Either way, we have shown that $q_1(1-q_1) \leq \dots \leq q_k(1-q_k)$ is in nondecreasing order, allowing the application of Chevychev's sum inequality above.

Putting Lemma~\ref{lem:lem2} together with the above results, as follows, yields the theorem.
\begin{align*}
\oE_{S \sim \mathcal{D}^n}\left[\oVar_{h \sim \ha}(\err_S(h))\right] &= (1 - 1/n)\cdot\oVar_{h \sim \ha}(\err(h)) \,+\, (1/n)\cdot\oE_{(x, y) \sim \mathcal D}\left[\oVar_{h \sim \ha}(\err_{x, y}(h)) \right] \\
&\geq (1/n)\cdot\oE_{x, y}\left[\oVar_h(\err_{x, y}(h)) \right] \\
&= \frac1n \oE_q\left[\sum_{c=1}^k q_c^2(1 - q_c)\right] \\
&\geq \frac{1}{nk} \oE_q\left[\sum_{c=1}^k q_c(1 - q_c)\right] \\
&= \frac{\err(\mathcal A)}{nk}.
\end{align*}
\end{proof}

\subsection{Replication of the main result from \citet{jiang2021assessing}}
Because it is theoretically related to our results, we include a simplified proof of the main result from \citet{jiang2021assessing}, which is Theorem 4.1 of that work.
\begin{theorem}[\citet{jiang2021assessing}]
If a stochastic training algorithm $\mathcal A$ is class-wise calibrated, then its error rate is equal to its expected disagreement rate between two trained networks.
\[
\err(\mathcal A) = \oE_{h_1, h_2 \sim \ha^2, (x, y) \sim \mathcal D}\left[1\{h_1(x) \neq h_2(x)\}\right]
\]
\end{theorem}
\begin{proof}
Let $q: \mathcal X \mapsto \mathbb [0, 1]^k$ be defined as in \Secref{app:proof4}. Then the laws of conditional expectation yield the following expression for the disagreement rate.
\begin{align*}
\oE_{h_1, h_2 \sim \ha^2, (x, y) \sim \mathcal D}\left[1\{h_1(x) \neq h_2(x)\}\right] &= \oE_q\left[\oE_{h_1, h_2, (x, y)}\left[1\{h_1(x) \neq h_2(x)\} \mid q(x) = q\right]\right] \\
&= \oE_q\left[\oE_{h_2, (x, y)}\left[\oE_{h_1}\left[1\{h_1(x) \neq h_2(x)\} \mid q \right] \mid q\right]\right] \\
&= \oE_q\left[\oE_{h_2, (x, y)}\left[\sum_{c=1}^k q_c \cdot 1\{h_2(x) \neq c\} \mid q\right]\right] \\
&= \oE_q\left[\oE_{x, y}\left[\oE_{h_2}\left[\sum_{c=1}^k q_c \cdot 1\{h_2(x) \neq c\} \mid q\right]\mid q\right]\right] \\
&= \oE_q\left[\oE_{x, y}\left[\sum_{c=1}^k q_c(1-q_c)\mid q\right]\right] \\
&= \oE_q\left[\sum_{c=1}^k q_c(1 - q_c)\right] \\
&= \err(\mathcal A).
\end{align*}
Each conversion of a conditional expectation over $\ha$ to a formula involving $q$ uses the assumption of class-wise calibration. The final step is via the fact that $\err(\mathcal A) = \oE_q\sum_{c=1}^k q_c(1 - q_c)$ as we showed in \Secref{app:proof4}.
\end{proof}

% \vspace{20mm}
\newpage

\begin{figure}
    \centering
    \includegraphics[width=0.98\textwidth]{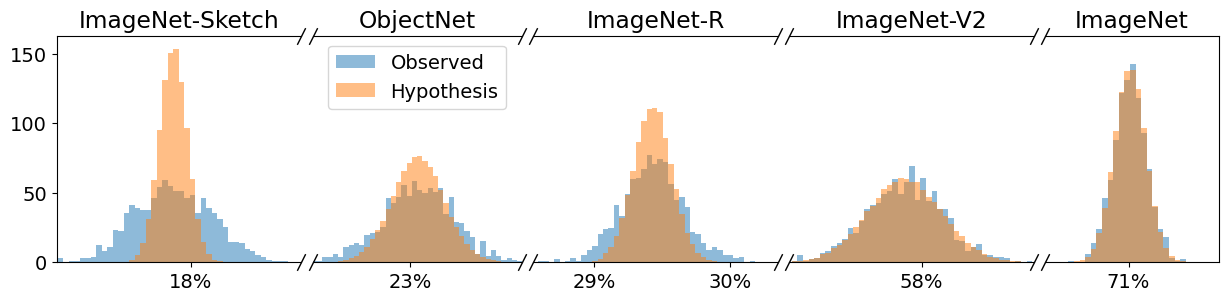}
    % \vspace{-0.1cm}
    \caption{\small \textbf{Distribution shift produces excess distribution-wise variance between runs.} Across 1,000 runs of ImageNet training, both the ImageNet validation set and ImageNet-V2 have accuracy distributions close to that predicted by the independent errors hypothesis, and hence, little distribution-wise variance. On the other hand, the accuracy distributions on distribution-shifted sets have significant excess variance, indicating genuine differences between trained models.}
    \label{fig:imagenet_shifts}
    % \vspace{-0.3cm}
\end{figure}

\section{ImageNet and distribution shifts}
\label{sec:imagenet}

In this section we show that shifted distributions of test data have increased variance in their accuracy distributions across repeated runs of training. We additionally confirm that the findings of \Secref{sec:variance} generalize to standard ImageNet training.

Our experiment is as follows. We independently train 1,000 ResNet-18s on ImageNet using a standard configuration (\Secref{app:hparams_in}). Their average top-1 accuracy is 71.0\%. We study the predictions of these networks on the ImageNet validation set, ImageNet-V2, and three shifted datasets.

We first look at the ImageNet validation set. In \Figref{fig:imagenet_shifts} (rightmost) we observe that the empirical accuracy distribution on this set closely matches the one predicted by the independent errors hypothesis. The observed standard deviation is 0.118\%, and \Eqref{eq:estimate_distributionwise} estimates that the distribution-wise standard deviation is 0.034\%. This value is close to what we found for CIFAR-10, confirming that both training scenarios adhere to the conclusions of \Secref{sec:variance}.

Next we consider ImageNet-V2~\citep{recht2019imagenet}. This dataset is intended to have the same distribution of examples as ImageNet, and we find that its accuracy distribution has similar statistical properties as well. In particular, we find that the distribution predicted by independent errors also closely matches its true distribution. \Eqref{eq:estimate_distributionwise} estimates a distribution-wise standard deviation of 0.071\%, which is larger than what we found on the ImageNet validation set, but still relatively small. We note that the test-set accuracy distribution for this dataset is wider, but this can be explained simply by the fact that it is $5\times$ smaller than the ImageNet validation set.

By contrast, ImageNet-R~\citep{hendrycks2021many}, ObjectNet~\citep{barbu2019objectnet} and ImageNet-Sketch~\citep{wang2019learning} all have different statistical behavior compared to the first two datasets. These datasets are constructed to have shifted distributions relative to ImageNet, and we find that their accuracy distributions have significant excess variance over that predicted by the independent errors hypothesis.
% We estimate using \Eqref{eq:estimate_distributionwise} that these three test-sets have large distribution-wise standard deviations of 0.181\%, 0.179\%, and 0.257\% respectively, indicating significant differences between runs of training.
We estimate using \Eqref{eq:estimate_distributionwise} that the distribution-wise standard deviations for these datasets are 0.181\%, 0.179\%, and 0.257\% respectively. This indicates significant genuine variability between repeated runs of training in terms of their performance on these distributions.

The above result provides empirical confirmation of a theory advanced by \citet{cohen2024ask}. The authors observed that pretraining provides a greater performance benefit for models evaluated on out-of-support distribution shifts like ImageNet-Sketch than it does for in-support shifts like ImageNet-V2. Seeking to understand this phenomenon, they analyzed a simple logistic regression setting and found that without pretraining, out-of-support shifts can induce more variance in behavior between runs of training than in-support shifts, due to a greater dependence on the initialization. Our above result, namely that ImageNet-Sketch has a distribution-wise standard deviation of 0.257\% across repeated ResNet-18 trainings whereas ImageNet-V2 has only 0.071\% ($13\times$ less variance), provides evidence that the logistic regression-based theory of \citet{cohen2024ask}, beyond being a source of useful intuition, is also directly true for neural networks.

\begin{figure}
    \centering
    \subfigure{\includegraphics[width=0.19\textwidth]{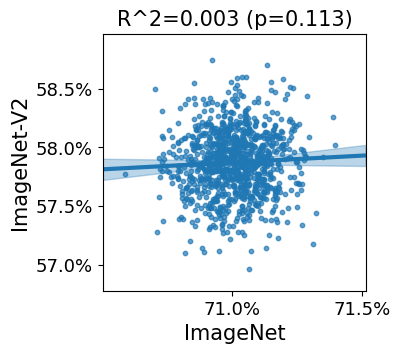}}
    \subfigure{\includegraphics[width=0.19\textwidth]{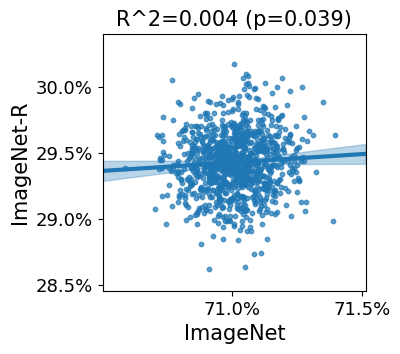}}
    \subfigure{\includegraphics[width=0.19\textwidth]{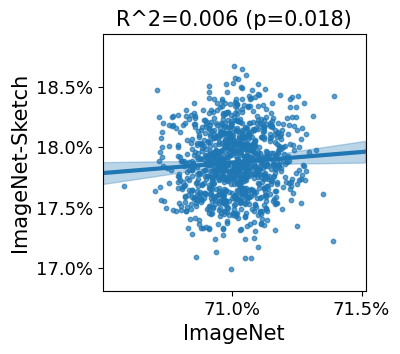}}
    \subfigure{\includegraphics[width=0.19\textwidth]{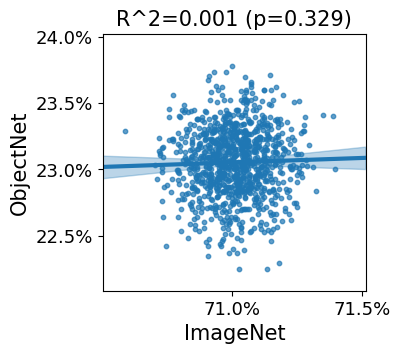}}
    \subfigure{\includegraphics[width=0.19\textwidth]{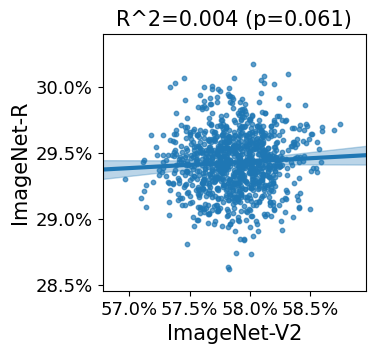}}
    \subfigure{\includegraphics[width=0.19\textwidth]{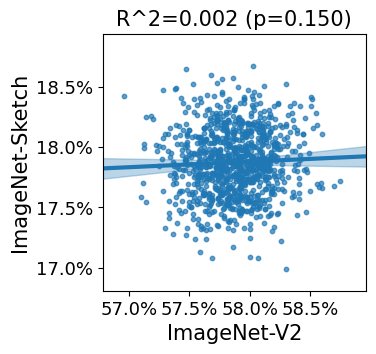}}
    \subfigure{\includegraphics[width=0.19\textwidth]{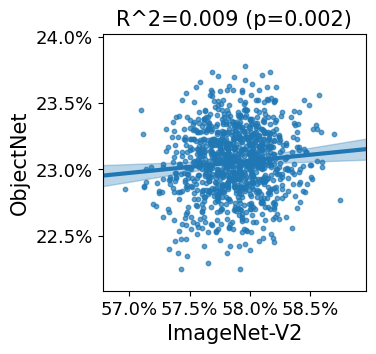}}
    \subfigure{\includegraphics[width=0.19\textwidth]{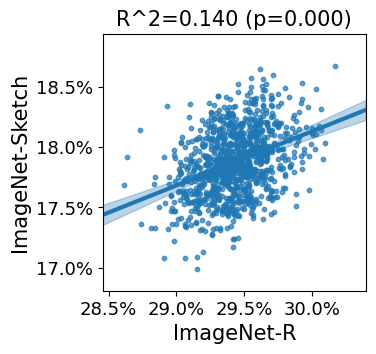}}
    \subfigure{\includegraphics[width=0.19\textwidth]{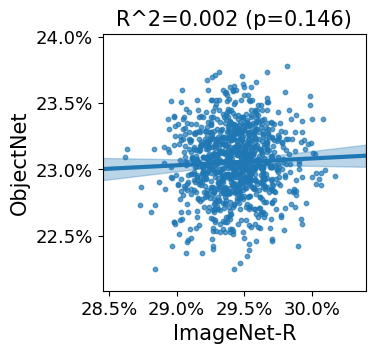}}
    \subfigure{\includegraphics[width=0.19\textwidth]{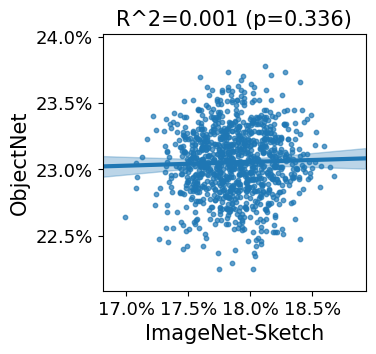}}
    \caption{\small \textbf{Correlations between distribution shifts.} We visualize the accuracy values of 1,000 ResNets which were independently trained on ImageNet. Each network is evaluated on the ImageNet validation set, as well as four extra datasets (IN-V2, IN-R, IN-Sketch, and ObjectNet). We display the scatterplots of accuracy on each of the $\binom{5}{2}$ pairs. The following pairs had statistically significant correlations: (ImageNet-R, ImageNet), (ImageNet-Sketch, ImageNet), (ObjectNet, ImageNet-V2), and (ImageNet-Sketch, ImageNet-R). All but one pair have weak correlations with $R^2 < 0.01$. The strong correlation is between ImageNet-R and ImageNet-Sketch with $R^2 = 0.14$. We hypothesize that this is caused by the fact that both sets contain many sketch-like images, so that this pair has a) similar distributions and b) shifted distributions relative to the training set. We report two-sided p-values.}
    \label{fig:distribution_shifts}
\end{figure}

In \Figref{fig:distribution_shifts} we additionally investigate correlations between pairs of these five datasets. The strongest correlation is between ImageNet-R and ImageNet-Sketch, with $R^2 = 0.14\,\,(p < 10^{-8})$. Manual inspection shows that both ImageNet-Sketch and ImageNet-R contain many sketch-like images, suggesting that similar features may induce correlation between distributions. All other pairs have $R^2 < 0.01$. For example, ImageNet-Sketch is decorrelated from ObjectNet, with $R^2 = 0.001\,\,(p = 0.34)$.

Overall, our findings suggest that training instability is in some sense a relative notion. ImageNet training is highly stable when evaluated on the main distribution, with a small standard deviation of 0.034\% on the underlying distribution of the ImageNet validation set.
But it is unstable on shifted distributions, with ImageNet-Sketch having a much larger standard deviation of 0.257\%.
This serves as a caveat to the main takeaway: from the perspective of the main training distribution, all runs perform at nearly the same level, but from the perspective of shifted distributions, there are sometimes significant differences between runs.

\section{Replication study of \citet{summers2021nondeterminism}}
\label{sec:factors}
\subsection{The three sources of randomness}
\label{sec:try_decompose}

Training neural networks typically involves three sources of stochasticity: model initialization, data ordering, and data augmentations.
In this section we investigate how each of these sources contributes to the final variance between runs that we observe at the end of training.

We develop a CIFAR-10 training framework\footnote{\url{https://github.com/KellerJordan/CIFAR10-isolated-rng}} that allows each source to be independently controlled by one of three different seeds. For example, when the data-augmentation seed is fixed and the data-order seed is varied, then the set of augmented images seen by the network throughout training will remain the same, but will be presented in a different order. When all three seeds are fixed, training is deterministic, so that repeated runs produce the same network every time. Standard training is equivalent to allowing all three seeds to vary.

Our experiment is to fix two seeds, and vary just the third (\eg varying only the data order while keeping the model initialization and data augmentations fixed). Our naive intuition is that each factor contributes some part to the overall variance,
so that this should decrease variance relative to the baseline of varying all three seeds.

\begin{figure}
    \centering
    \includegraphics[height=3.6cm]{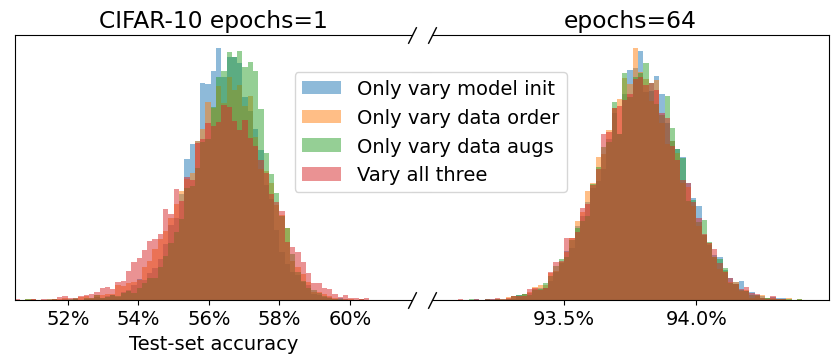}
    \includegraphics[height=3.6cm]{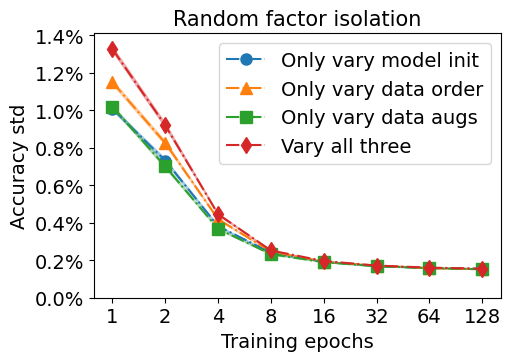}
    \caption{\small \textbf{One source of stochasticity yields the same variance as three for full training.} When training for only 1 epoch, varying all three sources of randomness induces a standard deviation of 1.33\% in test-set accuracy between runs, while varying any single source alone induces 25-40\% less variance. But when training for 64 epochs, varying any one source induces as much variance as all three together. Each distribution corresponds to 4,000 runs of training.}
    \label{fig:factors}
\end{figure}

We show the results in \Figref{fig:factors}. For short trainings of under 16 epochs, this intuition is correct. For example, when training for 4 epochs, if we fix the data order and augmentations, while varying only the model initialization, then variance in test-set accuracy is reduced by 26\%, such that the standard deviation decreases from $0.45\%$ to $0.38\%$.

However, for longer trainings of 32 epochs or more, varying just one of the three random factors produces approximately the same variance as the baseline of varying all three.
For example, across 4,000 runs of training for 64 epochs, varying just the model initialization (with data ordering and augmentation fixed) produces a standard deviation of 0.158\%, almost the same as the baseline, which has 0.160\%. At $n = 4,000$ runs of training this is not a statistically significant difference, so it is possible that the true values are the same, or that they differ by a small amount.
We conclude that for this training regime, any single random factor suffices to generate the full quantity of variance, rather than each factor contributing to overall variance.

\subsection{Sensitivity to initial conditions}
\label{sec:sensitivity}

\begin{figure}
    \centering
    \includegraphics[width=0.95\textwidth]{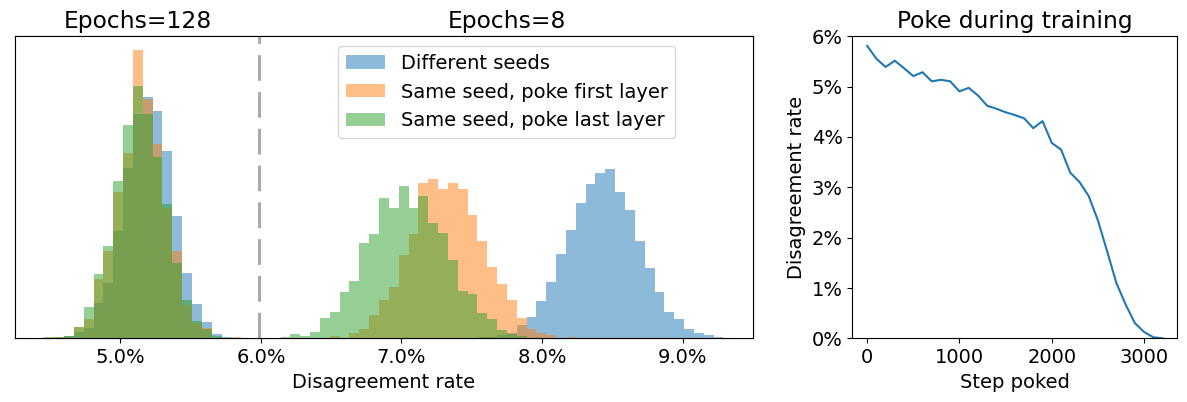}
    \caption{\small \textbf{Neural network training has high sensitivity to initial conditions.} (Left:) For short trainings, pairs of runs which differ only by one network having been ``poked'' (\ie had a single weight changed slightly at initialization) disagree on 7.0-7.5\% of predictions. Pairs of runs with fully different random seeds disagree more, on $\sim$8.5\% of predictions. For long trainings, there is almost no difference. The histograms are over repeated pairs of runs.
    (Right:) The earlier a network is poked during the training process, the more its predictions will disagree with the network that trained unperturbed from the same random seed.
    }
    \label{fig:poke}
\end{figure}

In the previous section, we showed that when training to convergence, varying just the model initialization (or just the data ordering, or augmentations) produces approximately the same quantity of variance between runs as a baseline fully random setup. In this section we find that even varying a single weight at initialization suffices.
Our findings replicate the work of \citet{summers2021nondeterminism}, who reach similar conclusions.

Consider multiplying a single random weight in the network by 1.001. We call this ``poking'' the network. This is a tiny change; recent work in quantization~(\eg \citealp{dettmers2022llm}) suggests that trained models can typically have \textit{all} their weights modified more than this without losing accuracy.

Nevertheless, in \Figref{fig:poke} we demonstrate that poking the network early in training produces a large difference in the final result. Our experiment is to run two trainings with the same random seed, but with one network being ``poked'' at some point during training. We measure the disagreement rate between the two networks, \ie the fraction of their test-set predictions that differ.
For short trainings, poking induces much less disagreement than changing the random seed. 
But when training for 128 epochs, poking alone produces an average disagreement of 5.14\%, barely less than the 5.19\% produced by using entirely different random seeds.
We have also observed that varying just the first batch of data, or the numerical precision of the first step (\eg fp16 vs. fp32) has a similar effect.
We conclude that almost all variation between runs is not produced by specific sources of randomness like model initialization, data ordering, etc., but is instead intrinsic to the training process, which has extreme sensitivity to initial conditions.

% \vspace{20cm}
% \newpage
\section{Additional figures}

\begin{figure}[H]
\begin{center}
    \includegraphics[width=0.5\textwidth]{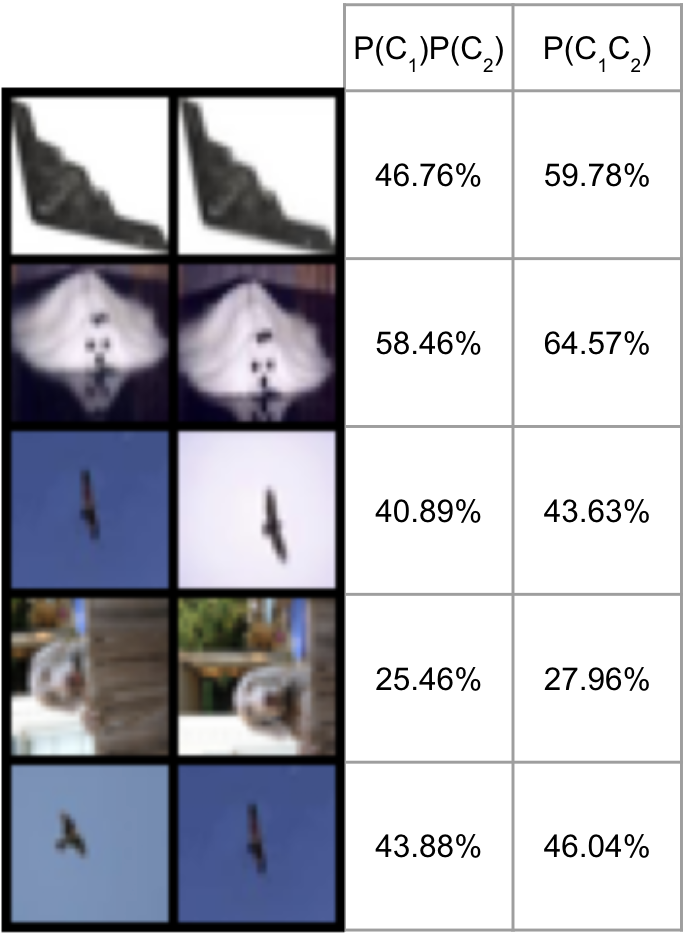}
    \caption{\small \textbf{There exist five pairs whose errors deviate by $\geq$2\% from independence.} The first column is the product of the probability (over training stochasticity) that the trained network predicts the first example correctly and the probability that it predicts the second example correctly. The second column is the probability that the trained network predicts both of them correctly. Each quantity is measured across 60,000 runs of our 64-epoch training configuration. The independent errors hypothesis (Definition~\ref{def:independent}) predicts that these two quantities should be equal. Out of all $\binom{10,000}{2}$ pairs of examples in the CIFAR-10 test-set, only these five deviate by more than 2\% from that prediction. The remaining 49,994,995 pairs are all within 2\% of that prediction.}
    %, \ie have $|P(C_1C_2) - P(C_1)P(C_2)| \geq 0.02$
    \label{fig:anompair}
\end{center}
\end{figure}

\begin{figure}
    \centering
    \hspace{1.5cm}\includegraphics[width=0.6\textwidth]{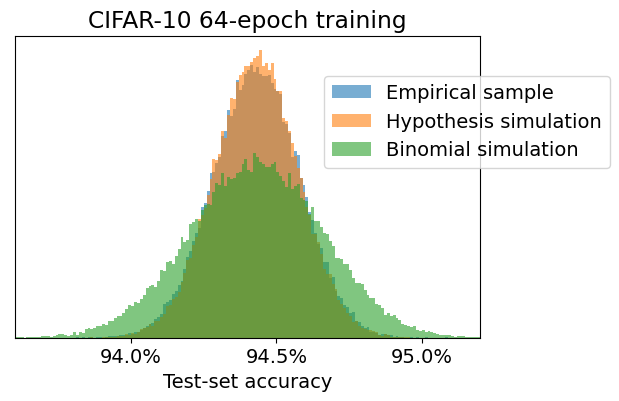}
    % \hspace{2cm}\includegraphics[width=0.5\textwidth]{figures/binomial.png}
    \caption{\small \textbf{The binomial approximation overestimates variance}. Compared to the empirical distribution of test-set accuracy, the binomial approximation predicts a distribution with too much variance. We use $p = 0.9441$ (the average accuracy) and $n = 10,000$ (the size of the test-set) to simulate $60,000$ samples from $\mathrm{Binom}(n, p)$, which we find overestimates variance by a factor of $\approx 2.5\times$. In comparison, the framework of independent errors (Definition~\ref{def:independent}) provides an accurate estimate.}
    \label{fig:binomial}
\end{figure}

\end{document}